# CAE-LO: LiDAR Odometry Leveraging Fully Unsupervised Convolutional Auto-Encoder Based Interest Point Detection and Feature Description

Deyu Yin [1, 2], Qian Zhang [3], Jingbin Liu [1, 2, *], Xinlian Liang [2], Yunsheng Wang [2],
Jyri Maanpää [2], Hao Ma [1], Juha Hyyppä [2], Ruizhi Chen [1]

[1] State Key Laboratory of Information Engineering in Surveying, Mapping and Remote Sensing, Wuhan University, China
[2] Department of Remote Sensing and Photogrammetry and the Center of Excellence in Laser Scanning Research, Finnish Geospatial Research Institute, Finland
[3] School of Economics and Management, Hubei University of Technology, China
* jingbin.liu@whu.edu.cn

## Abstract

*As an important technology in 3D mapping, autonomous driving, and robot navigation, LiDAR odometry is still a challenging task. Appropriate data structure and unsupervised deep learning are the keys to achieve an easy adjusted LiDAR odometry solution with high performance. Utilizing compact 2D structured spherical ring projection model and voxel model which preserves the original shape of input data, we propose a fully unsupervised Convolutional Auto-Encoder based LiDAR Odometry (CAE-LO) that detects interest points from spherical ring data using 2D CAE and extracts features from multi-resolution voxel model using 3D CAE. We make several key contributions: 1) experiments based on KITTI dataset show that our interest points can capture more local details to improve the matching success rate on unstructured scenarios and our features outperform state-of-the-art by more than 50% in matching inlier ratio; 2) besides, we also propose a keyframe selection method based on matching pairs transferring, an odometry refinement method for keyframes based on extended interest points from spherical rings, and a backward pose update method. The odometry refinement experiments verify the proposed ideas' feasibility and effectiveness.*

## 1. Introduction

LiDAR Odometry (LO) plays an important role in autonomous driving [1], robot navigation [2], indoor mapping [3, 4], and outdoor mapping [5], etc. But until now it's still a challenging task and the main reasons come from the high accuracy demand and the difficulties on the processing of LiDAR data. Firstly, LiDAR odometry is sensitive on error accumulation especially on the rotation error at a long distance. Secondly, it is difficult to extract information from LiDAR point cloud due to its unorder and a great amount of points.

Before the revolution of deep learning, LiDAR odometry solutions were based on handcrafted algorithms. But this is usually complicated and the parameters are difficult to adjust manually. The deep learning technology is designed to learn parameters automatically, which means it can greatly decrease the cost of development and the complexity of the algorithms. Recently, some key technologies based on deep learning that can be used in LiDAR odometry [6-8] have emerged.

But the input data structure is still a very important problem regardless of whether the algorithm is based on deep learning or not. The large amount of unordered points is the first problem to face. Although PointNet [9] is proposed to solve this problem, but it has the limitation on the number of input points [10]. From another way, projecting point cloud into grid structured data is a common approach to order the points. Also, grid structured data let the use of more mature Convolutional Neural Networks (CNNs) be very suitable and convenient.

The choice of what kind of structured data to use depends on the application. Generally, there are two traditional ways to project LiDAR point cloud into structured data: 2D grid and 3D grid. These two structures can be also called image model and voxel model. 2D grid is a more compact way while 3D grid is scale isotropy. Interest point (or keypoint) detection and feature description are two key technologies in LiDAR odometry. The interest point detection part is to pick interest points out from the whole input point cloud, and this indicates that it needs efficient computing from the compact 2D grid. However, the feature description part is to describe the features corresponding to the detected interest points using their neighbor data, which needs the merits of preserving original 3D shape from 3D grid, despite its sparsity. Hence, we use 2D grid for interest point detection and use 3D grid for feature description.

Another important aspect for deep-learning-based methods is the supervision type. There are several methods that use weakly supervised way, such as taking ground-truth poses as a kind of constraint to train the networks [6, 11]. While in our method, Convolutional Auto-Encoders (CAEs) are applied to achieve fully unsupervised training, which has a higher universality.

Besides these two key technologies, new methods for the odometry refinement part in LiDAR odometry are also proposed. In total, the proposed method is named as Convolutional Auto-Encoder based LiDAR Odometry (CAE-LO) and the main contributions are:

- 2D-CAE-based interest point detection method on spherical ring LiDAR point cloud.
- 3D-CAE-based multi-scale feature extraction method on voxel model.
- Keyframe selection method based on feature matching pairs transferring.
- Pose transformation backward update method between keyframes.

Our code is available online[1].

## 2. Related Work

The amount of literature related to this work is huge. Since we follow the feature-based frame to frame matching pipeline for the generation of LO, related works always revolve interest point detection and feature description. As is said in the former section, the point cloud data structure is a key problem. This section discusses the methods based on different point cloud representation, interest point detection methods, and feature extraction methods.

### 2.1 Data Representation Methods of LiDAR Point Cloud

Currently, the data representation methods of LiDAR point cloud for extracting information can be divided into three subcategories: unordered points，2D grid, and 3D grid.

PointNet [9] and PointNet++ [12] were proposed aiming at solving the disordering input problem of raw point cloud. They and many PointNet-like networks are used for doing 3D classification, segmentation, and hierarchical feature learning. Also, many PointNet-based networks emerged to detect interest points and to extract features from LiDAR point cloud, such as 3DFeat-Net [6], L3-Net [13], USIP [8], DeepICP [14], VoxelNet [15], etc. This way of using unordered points as input can save the information from the raw data as much as possible. However, due to the limitation on the number of input points, many methods need to downsample the point cloud in advance, which means the information is lost inevitably. From the aspect of extracting features, this can lead to a low feature description ability. For example, there are only 64 neighbor points near to a detected interest point in [6, 13] and only 35 sampled points within the same big-size voxel in [15] are used for feature description. Therefore, a fusion of local features is needed to improve the feature description ability [8, 11, 12].

---
[1] https://github.com/SRainGit/CAE-LO

2D grid representation has gained the most attention because it offers a good trade-off between complexity and detection. The most general projection way is to put the points into a 2D array along with a specific direction, such as Watertight [16]，Bird's Eye View [17, 18] or Front View of LiDAR point cloud [19]. But they lose some information if there are overlaps along the projection direction, and they are also not compact enough for LiDAR point cloud. For the multi-beam LiDAR, according to its scanning geometry, the point cloud can be transformed into a 2D grid via cylindrical [20] or spherical [21] projections. This is a very compact LiDAR point cloud projection method with a relatively low information loss and the projection content can be various types such as 3D coordinate values, intensity information, or range information to form a multi-channel 2D data [7, 20, 22-24]. So, it is very suitable to use 2D CNN, and its effectiveness has been shown in the applications such as vehicles detection [20], ground segmentation [24], object segmentation [25], semantic segmentation [22], and even end-to-end LiDAR point cloud scan-to-scan matching [7, 23]. However, its shortcoming is the scale difference in different pixel locations. The shared parameters of fixed-size filters in CNN will get different receptive field sizes in the real 3D world when computing on pixel patches where the distances to the LiDAR sensor are different. Hence, trying to utilize its advantages and avoid its shortcomings simultaneously, we decided to use 2D CNN on 2D spherical ring (details are in Section 3.2) to detect interest points.

Unlike 2D grid, 3D grid, which is also called as voxel model, has a constant scale in real size on every voxel inside it. It also preserves the original shape of the input point cloud. For example, in [20], a 3D fully convolutional network is applied to detect vehicles using voxel model. But it is computationally expensive when using voxel models on a whole LiDAR point cloud voxel model. Most of the voxels are empty and the computation is much higher than 2D grid due to its 3D structure. The memory cost is very high if the voxel size is small with the purpose of improving data resolution. In order to still take advantages of voxel model, using only voxel patches to extract features can minimize its disadvantages. In [26, 27], the voxel patches surrounding to keypoints are used to extract features, and the voxel sizes are 0.02m and 0.01m respectively. Similar to these approaches, our multi-scale features are based on voxel patches, but with multiple resolutions.

## 2.2 Interest Point Detection

Interest point detection is to select matchable point candidates from the whole data. Traditional handcrafted interest point detection algorithms are ISS [28], Harris 3D [29], SHOT [30], NARF [31], and clustering method used on watertight model [16], etc.. After the rise of deep learning, there are methods that combine handcrafted operators with deep learning [32] and the methods fully based on deep learning. As far as we know, USIP is state-of-the-art to detect interest points from raw point cloud. Even so, this method still cannot find out satisfied interest points in some scenarios.

The deep learning based methods can be divided into three categories: supervision-based methods, weakly-supervised methods, and unsupervised methods. Supervision-based methods use classifiers [33-36] or end-to-end approach [37] to recognize keypoints. Weakly supervised methods usually use the constraint from ground truth pose [6, 38, 39] or feature points generated by Structure from Motion (SfM) [40] to train the networks to detect interest points. There are a few works based on unsupervised learning. USIP [8], which takes raw points as input, is one of them.

The work in [41] creatively leverages L2-norm of the output of CNN filters to re-localize interest points in images. Our method was inspired by this idea. But different from it, our method directly uses the output of CNN filters as a kind of local feature vector to detect interest points.

## 2.3 Feature Description

Feature description refers to describe the input data using a d-dimensional vector in a concise way and it also can be interpreted as a kind of data compression. Regardless of global descriptors, no matter the handcrafted descriptors or deep-learning-based descriptors, the feature description is based on an area surrounding to the interest point.

In order to get features with high matchability, the scale that feature can describe is an important factor. Although some methods use Siamese network [40], triplet loss [35, 37], or even N-tuple loss [11] to train the descriptor to improve the matchability, the scale is always the bottleneck. Generally speaking, if the resolution of the data is fixed, then the bigger the scale results in more matchability, while, at the same time, the more hardware resources are needed. For the methods to take raw points as input, they have to use hierarchical strategy [12, 42] or feature fusion methods [11, 43] to strengthen the matchability. In the domain of 3D-grid-based [26, 27] or 2D-image-based [35, 37, 40] feature matching, in which the area that feature describes is a 3D/2D patch, a fixed scale with a fixed resolution always has a limited matchability. To conquer these problems, a fully unsupervised 3D descriptor with multi-scale voxel patch is proposed in this paper.

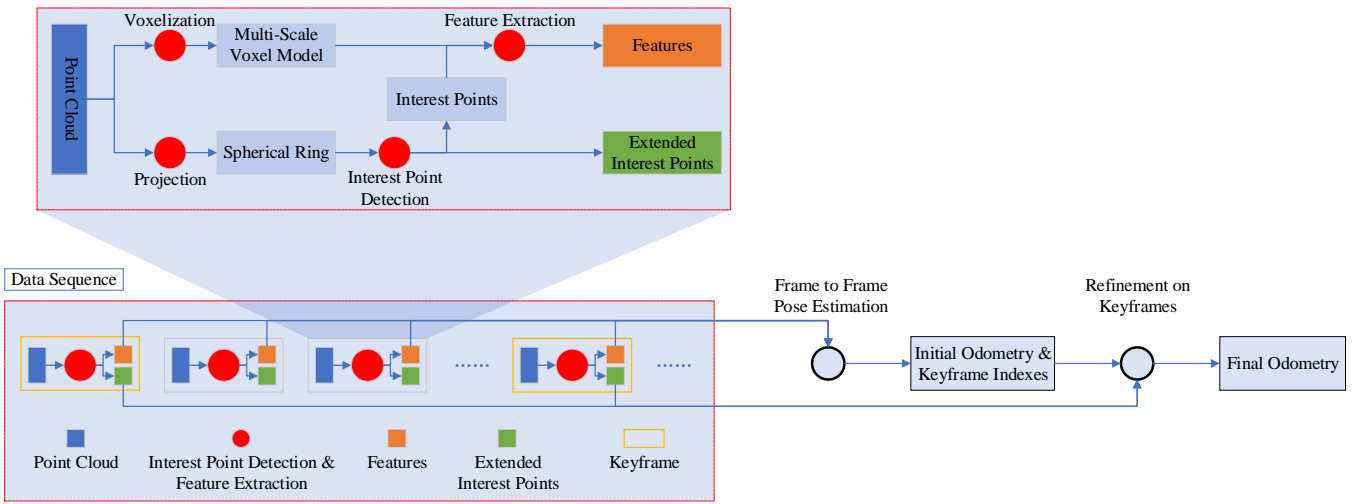

**Figure 1.** Overview of the method.

## 3. The Proposed Method

### 3.1 Outline of the Method

The method contains two main steps. The first step takes point cloud sequence as input and applies frame to frame pose estimation to generate initial odometry and also to pick out keyframes. The second step is an ICP-based odometry refinement which only applies to keyframes to get the final refined odometry.

As shown in Figure 1, the initial odometry is generated from the frame to frame pose estimation based on the extracted features corresponding to the detected interest points. The generation of Extended Interest Points (EIPs), which is used for the odometry refinement on keyframes, is based on the interest point detection. Besides these, the supporting methods of keyframe selection method and backward pose update method are also elaborately designed and included in the odometry refinement part.

In our method, the interest point detection from spherical ring and the feature extraction from multi-scale voxel model are all based on convolutional auto-encoder. See below for details.

### 3.2 Unsupervised Interest Points Detection

Spherical ring is a very compact way to represent LiDAR point cloud data. On the one hand, it's a 2D matrix data representation that is suitable for convolutional operations. On the other hand, due to its compactness, there are few empty pixels which can lead to useless computations.

Different from some methods [20, 22, 25], which use only the front part of LiDAR point cloud, we project the whole LiDAR point cloud into a spherical ring, trying to take full use of all the information. Assume there is one point in the point cloud $p = (x, y, z)$, and the projection functions are:

$$c = (\pi - arctan2(y, x))/\Delta\alpha \quad (1)$$
$$r = H - \left(\arcsin(z/\sqrt{x^2 + y^2 + z^2})/\Delta\beta - \beta_d/\Delta\beta\right) \quad (2)$$

where $c$ and $r$ represent the row number and the column number of the grid cell to project in respectively, $H$ is the height of the spherical ring, $\Delta\alpha$ and $\Delta\beta$ are the angle resolutions for the laser beams in horizontal and vertical direction respectively. $\beta_d$ is the pitch angle of the lower laser beam and the item of $-\beta_d/\Delta\beta$ is to make sure that all the projected image coordinates are positive values.

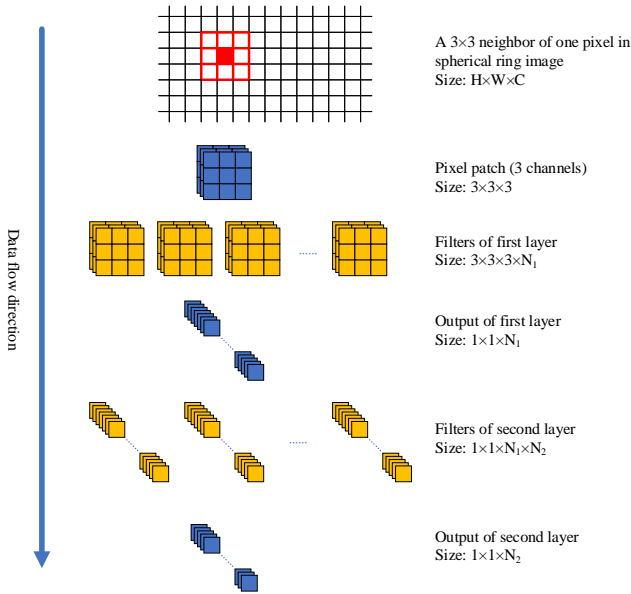

**Figure 2.** The basic idea of a $3 \times 3$ neighbor feature description using convolutional filters. $N_1$ is the number of first layer filters. $N_2$ is the number of second layer's filters.

With the purpose of making the method more universal and concise, we put the coordinate values $(x, y, z)$ into the spherical ring model even though the range and intensity data are available in the dataset is going to be used. To the end, the spherical ring model will be a 2D matrix with a dimension of $H \times W \times C$. Here $W$ is the width of the spherical ring, and $C$ is the number of channels.

Taking the 2D structured spherical ring data as input, we address the problem of interest point detection as picking out the pixels corresponding interest points that have big difference values with their neighbor pixels. To measure the difference value, we use the output of CNN filters as the feature description of the local pixel area which is also the receptive field of the filters, and then the L2 norm of the difference between the features is taken as the difference value.

Figure 2 depicts this basic idea. Once the training of CNN is finished, the parameters in the filters (showed with yellow grids) are fixed. The output of one layer of filters for one pixel of input will be a vector, which will be taken as the feature descriptor of the corresponding receptive field.

To achieve the goal of computing the features for all the pixels in spherical ring with an unsupervised training way, we propose to use CAE. Different from Auto-Encoders (AEs), the parameters in CAEs are shared, which means the weights are shared among all locations in the input, preserving spatial locality [44]. Once the network is trained in this convolutional manner, the whole feature map can be obtained with the whole spherical ring as input. In total, this method has the following merits: 1) it works in an end-to-end manner; 2) the network is trained in an unsupervised manner; 3) the network is very light; 4) the network is easy to train. The overview of training and inference is shown in Figure 3.

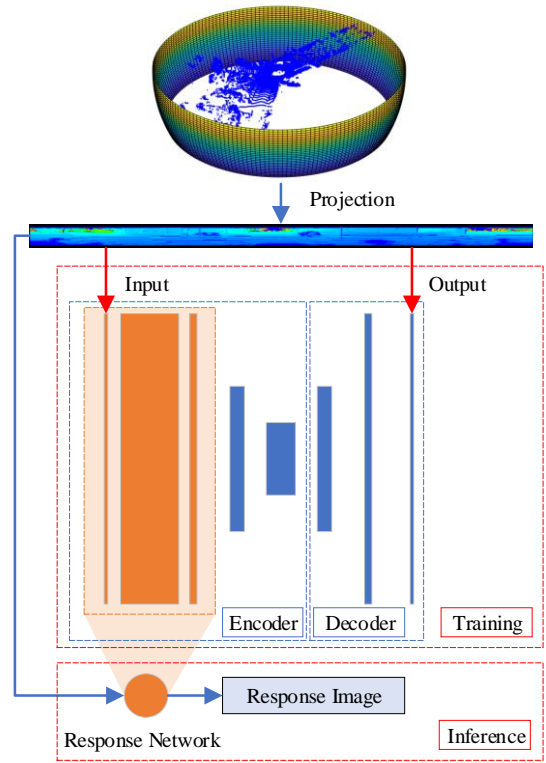

**Figure 3.** Overview of the training and inference of response image convolutional neural network.

The bottle-neck structure in which the output size of the layers first decreases and then increases is a typical feature of AE networks. The purpose of this feature is to force the network training filters to perform data compression. Meanwhile, on the contrary, the number of channels generally increases first and then decreases. But the number of channels in the second layer is bigger than in the third layer, aiming to increase the non-linearity for enhancing the ability to describe the local feature, and simultaneously decrease the dimension of the response output to save computational effort.

The meta structure of the response network corresponds to Figure 2. The third layer in Figure 3, which is shown as the second convolutional layer in Figure 2, uses $1 \times 1$ convolution. Thus, the receptive field of the response layer for each pixel is still $3 \times 3$. Assume the input spherical ring has a size of $H \times W \times C$. Once it passes the response network, the response image $R$ can be obtained, and the size is $H \times W \times N_2$, where $N_2$ is the dimension of the features. To get the differences with neighbor pixels, the following procedures should be carried out.

Assume the neighboring area is $(2h+1) \times (2h+1)$, then each pixel has $((2h+1)^2 - 1)$ pixels to compare. Firstly, we record all the response differences with neighbor pixels for each pixel. Secondly, still for each pixel, by using the mask generated by the projection procedure, the smallest difference among the valid neighbors is picked out as its score. Thirdly, all the scores in the whole spherical ring are ranked and the points in the pixels that have high scores are taken as interest points.

The computing of the feature difference map follows:
$$D(r,c,u,v,:) = R(r,c,:) - R(r+u, c+v,:) \quad (3)$$
where $D$ is the feature difference map with a size of $H \times W \times h \times h \times N_2$. There are $h \leq r < H-h$, $h \leq c < W-h$, $-h \leq u \leq h$, and $-h \leq v \leq h$. And to quantify the difference, an L2-norm is applied to the feature difference map to get the difference map $D_N$. The computation follows:
$$D_N(r,c,u,v) = \mathcal{N}\big(D(r,c,u,v,:)\big) \quad (4)$$
where $\mathcal{N}$ is a function to compute the L2-norm of the input vector. So, the purpose of $D_N$ is to record all the difference values of the neighbor pixels for each pixel. The qualification of the score for being an interest point is to take the smallest difference among its valid neighbors. The computation of the score map with the mask for available pixels follows:
$$S(r,c) = Min\big(D_N(r,c,:,:), \mathcal{M}(r-h:r+h, c-h:c+h)\big) \quad (5)$$
where $\mathcal{M}$ is the mask which indicates the valid pixels in the spherical ring, and its size is $H \times W \times 1$. The function $Min$ is to get the smallest valid value in the input matrix $D_N(r,c,:,:)$ with its mask, which size is $(2h+1) \times (2h+1) \times 1$. Hence, the score map, which size is $H \times W \times 1$, describes the difference between each pixel in the spherical ring and its surrounding pixels to score it as an interest point.

To get not too many interest points, a threshold $\delta$ is set to filter out mediocre points and a limitation of $N_\delta$ is set as the maximum number of interest points to get. To avoid getting too many interest points from a near distance, we set a threshold $\sigma$ as the nearest distance to be accepted as an interest point. After this, EIPs can be obtained by picking out valid neighbor points in the spherical ring within a neighbor of size $h_E$.

### 3.3 Multi-Scale Feature Extraction

As mentioned in Section 2, a local feature is a vector to describe a local area of data. Thereby, the size of the region to which the extracted features correspond is a key factor in feature matching performance. However, at the same time, the matching accuracy should be also considered, which is determined by the resolution of data representation. In general, the larger the area that the feature can correspond to and the higher the resolution of data representation, the higher the matching stability and the higher matching accuracy accompanied by higher computational load.

A local feature based on voxel model is extracted from a local voxel patch centered on an interest point. Based on this, to achieve high matching stability and high matching accuracy, we propose to use a multiple resolution voxel models and a fixed size of voxel patches to extract features. As illustrated in Figure 4, assume the patch size is $S_P$, which means there are $S_P \times S_P \times S_P$ voxels in one patch. Set $S = \{S_1, S_2, S_3\} (S_1 < S_2 < S_3)$ are the voxel sizes. Since the patch size is fixed, the size of the covered area by the patch increases from $S_1$ to $S_3$, while the computational load stays the same. Here we set $S_2 = 8 \times S_1$ and $S_3 = 32 \times S_1$. Also, we set $S_1 = 0.02m$ and $S_P = 16$, then the corresponding area of patches based on resolutions set $S$ will be $\{0.32m, 2.56m, 10.24m\}$. Such a big variance of the multiple scales is an insurance of the feature description ability.

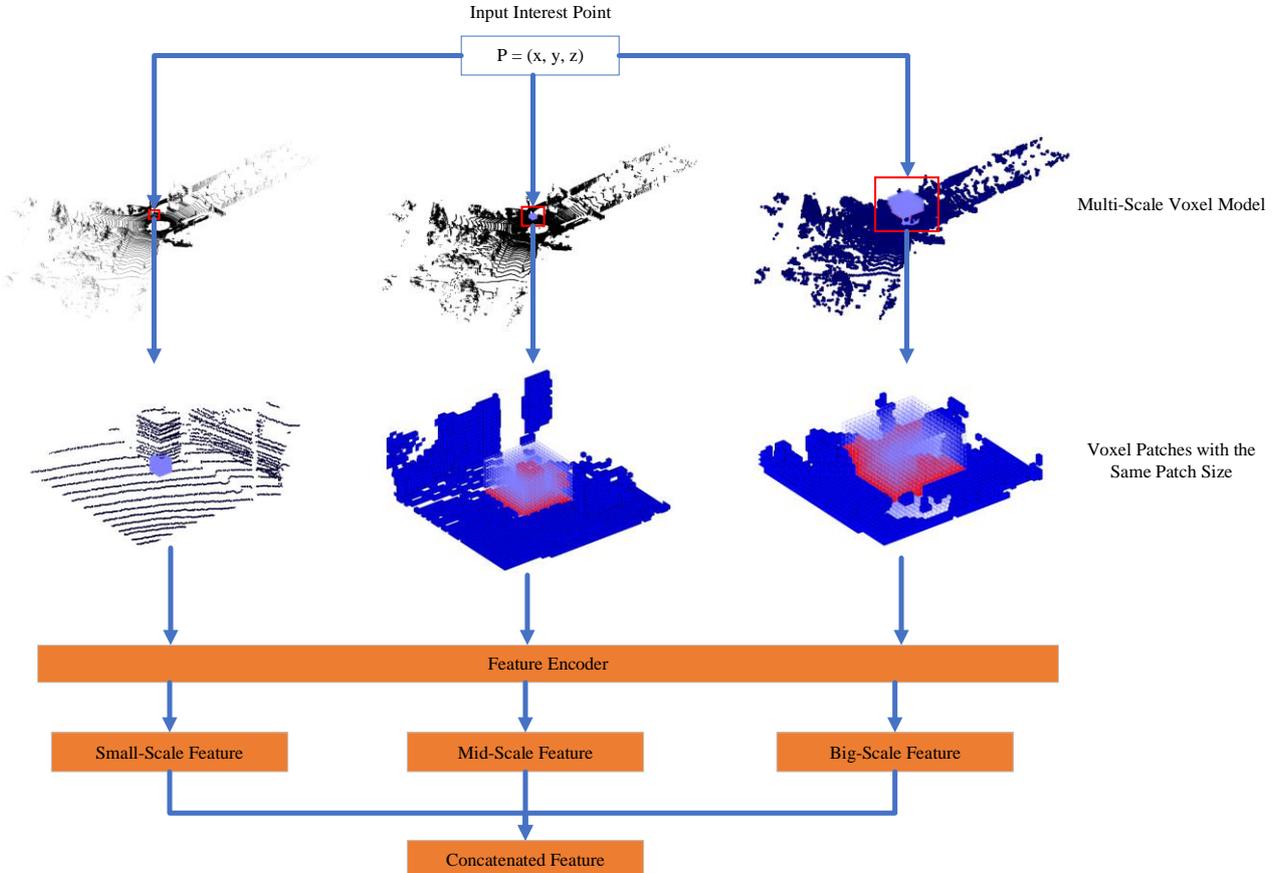

**Figure 4.** The process of obtaining multi-scale features from a 3D point. The voxels are represented as blue cubes. Red voxels are the valid voxels in voxel patches. Blue grids are the outlines of voxel patches.

Setting a fixed patch size on all the resolutions also allows them to be input into the same CAE network to achieve parameter sharing. Therefore the obtaining of multi-scale feature can be done by concatenating the three outputs together from the same network.

In [27], the voxel patch is input into the 3D CNN network to extract a feature vector. The difference to their method is that the training of our network is fully unsupervised. The same applies with the 2D CAE for interest point detection, the input and the output are the same to force the network to compress the input data and to decompress it back during the training of 3D CAE.

As shown in Figure 5, the 3D CAE contains two parts: encoder and decoder. The training in the 2D CAE for detecting interest point is performed to obtain the response layer, while the training in this 3D CAE is done for obtaining the compressed code as the feature. $S_P = 16$ is a canonical patch size. We use pooling layers and fully connected layers to compress the data. After the training, one 20-dimensional vector can be computed from each input voxel patch. Three voxel patches can be extracted according to the coordinate values of each interest point in three different scales. Finally, the three 20-dimensional vectors can be concatenated to a 60-dimensional vector to be the multi-scale feature of the interest point.

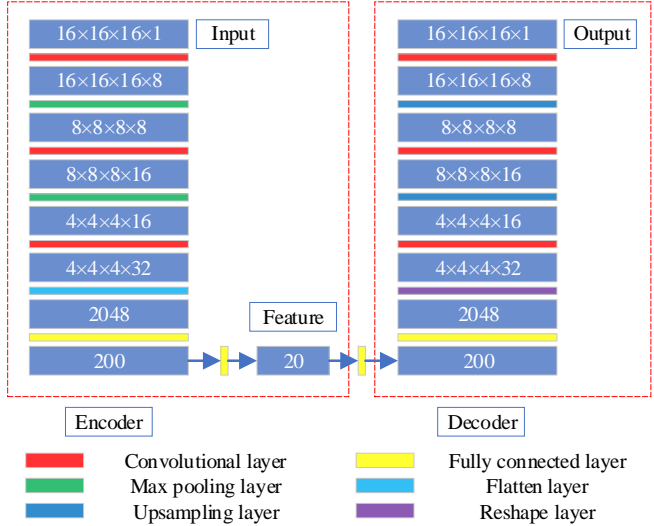

**Figure 5.** Flow chart of the output sizes of the layers in the 3D convolutional auto-encoder.

In general, there are two ways to prepare the training dataset. Either way is based on selecting a point and then extracting the corresponding three voxel patches. One way to do this is by picking up the points randomly from the raw point cloud and another way is by picking up the points randomly from the detected interest points. In order to make the feature description more specific, the latter way is chosen. In addition, because $S_P$ is an even number in our case, we use the integers in the close zone $\left[x_V - \frac{S_P}{2}, x_V + \frac{S_P}{2} - 1\right]$ ($x_V$ is one of the integer coordinate values among the three dimension directions) as the coordinate indexes of the voxels.

### 3.4 Keyframe Selection and Odometry Refinement

The initial odometry can be got by using RANSAC based frame to frame feature matching. To get a more accurate odometry, a refinement for keyframes is carried out.

The selection of keyframes should follow some metrics. Selecting keyframes by distance is a canonical way. But this method is not reliable if the initial odometry has a low accuracy. Especially if there are multiple consecutive failed matchings, the distance between two keyframes might be much bigger than in reality, leading to a small overlap or even no overlap between them.

To ensure sufficient overlap between keyframes, we utilize the matching pairs transferring during the frame to frame matching to select keyframes. Thus, the length between two keyframes will be limited by the transferring of feature matching pairs, not too short or too long.

Assume there are $n_{I_m}$ interest points with features in the $m_{th}$ frame, and we set $I_m = \{1,2,\cdots n_{I_m}\}$ as the index set of the points. We denote $p_k^m = \langle i_k^m, i_k^{m+1}\rangle$ ($i_k^m \in I_m, i_k^{m+1} \in I_{m+1}$) as one matching pair between frame $m$ and frame $m + 1$. Then we take $P_m = \{p_1^m, p_2^m, \cdots p_{P_m}^m\}$ as the set of matching pairs. Here, a function $f$ is defined to find the corresponding point indexes in frame $m + 1$ according to the input point indexes in frame $m$:

$$I'_{m+1} = f(P_m, I'_m) \ (I'_m \subset I_m, I'_{m+1} \subset I_{m+1}) \quad (6)$$

For example, according to the definition, we can get the set of the point indexes transferred from frame $m$ to frame $m + 1$. If the set is empty, then it means the transfer is failed. The keyframe selection can be started from taking the first frame in the frame sequence as the first keyframe. The previous frame of the frame that the first failed transferring will be taken as the second keyframe. Then the third keyframe can be found by starting the transferring from the second keyframe. Repeat this process until all the keyframes are found.

The pose refinement for the keyframes is based on ICP [45]. However, considering that it will be slow if we put the whole LiDAR point cloud into the registration, the use of a much smaller number of EIPs reduces the computational load. Since the detected interest points will be always in "structured" locations according to the detection principles, the registration will be less affected by meaningless areas.

Assume $A$ and $B$ are the extended interest point clouds in the previous keyframe and in the current keyframe respectively. B is transformed into the coordinate system of $A$ in advance. The registration is to find a pair of optimized rotation matrix $\boldsymbol{R}$ and translation vector $\boldsymbol{t}$. The optimization function is:

$$\arg\min_{R,t} \sum_{i=1}^{|A|}\sum_{j=1}^{|B|} w_{i,j}\|\boldsymbol{a}_i - (\boldsymbol{R}\boldsymbol{b}_j + \boldsymbol{t})\|^2,$$
$$w_{i,j} = \begin{cases} 1, \boldsymbol{a}_i \text{ corresponds to } \boldsymbol{b}_j \\ 0, \text{otherwise} \end{cases} \quad (7)$$

where $\boldsymbol{a}_i \in A, \boldsymbol{b}_j \in B$.

During the experiments, we found that the results are not accurate enough if we put all the EIPs into all the iterations of ICP. Therefore, we use a threshold to reject correspondence pairs in which the point-to-point distance exceeds it. This threshold decays exponentially during the iterations. The threshold is large at the beginning so that the objective function converges quickly. Whereas the latter iterations with smaller thresholds lead to a fine-tuning registration result.

The refinement of keyframes can reduce the accumulated errors significantly. However, it only updates the relative poses of keyframes not the relative poses of the ordinary frames between the keyframes. This leads to two problems: 1) Firstly, it will lead to a pose jump between the keyframes and the previous frames of the keyframes, resulting in an unreal trajectory of the moving platform. 2) Secondly, the accuracy of the ordinary frames' poses is lower than it should be. So, a backward pose update method for the ordinary frames' poses is carried out.

The basic idea of the backward pose update is that the change of the keyframe's pose after the refinement is evenly distributed to each ordinary frame between keyframes. Firstly, denote the rotation and translation matrix as equation (8). Assume the frame indexes of the two keyframes are $0$ and $n$, and there are $n - 1$ frames to compute pose update. As for the frame $i(i \in [1, n-1])$, based on the original pose, the update pose $\Delta \boldsymbol{T}_i$ should satisfy equation (9).

$$\boldsymbol{T} = \begin{bmatrix} \boldsymbol{R} & \boldsymbol{t} \\ 0 & 1 \end{bmatrix} \quad (8)$$

$$\boldsymbol{T}_{1,i}\Delta\boldsymbol{T}_i\boldsymbol{T}_{i+1,n} = \Delta\boldsymbol{T}_i^*\boldsymbol{T}_{0,n}^{\text{Ori}} \quad (9)$$

where $\boldsymbol{T}_{1,i}$ and $\boldsymbol{T}_{i+1,n}$ are the accumulated pose transformations from frame $0$ to frame $i$ and from frame $i$ to frame $n$ respectively. The computation is performed in the following equation:

$$\boldsymbol{T}_{j,k} = \prod_{i=j}^{k}\boldsymbol{T}_i \quad (10)$$

After the refinement, denote the change of the pose at frame $n$ as $\Delta\boldsymbol{T}^*$. The solving of $\Delta\boldsymbol{T}_i$ should be performed stepwise from $1$ to $n - 1$ according to equation (9). $\Delta\boldsymbol{T}_i^*\boldsymbol{T}_{0,n}^{\text{Ori}}$ can be taken as the updated pose of frame $n$ during the solving of $\Delta T_i$. Each pose update of the frames from frame 1 to frame $n-1$ leads around $\frac{1}{n}\Delta T^*$ pose change for frame n. After the $i_{th}$ update, the total pose change of frame $n$ is $\Delta T_i^*$, as in equation (11). $T_{0,n}^{\text{Ori}}$ is the accumulated pose translation before the refinement from frame 0 to frame $n$, as in equation (12).

$$\Delta T_i^* \approx \frac{i}{n}\Delta T^* \tag{11}$$

$$T_{0,n}^{\text{Ori}} = \prod_{j=1}^{n} T_j^{\text{Ori}} \tag{12}$$

To achieve the computation of equation (11), $\Delta T_i^*$ is divided into two parts $\Delta R_i^*$ and $\Delta t_i^*$. The computation of $\Delta T_i^*$ is shown in equation (14). The solving of the rotation matrix is non-linear computing, and it is computed by equation (13). Function $R2Eulers$ is to convert rotation matrix into three rotation Euler angles, and function $Eulers2R$ is to convert the Euler angles into a rotation matrix. Among the two convertings, the rotation order of the three axes follows "XYZ".

$$\Delta R_i^* = Eulers2R\left(\frac{i}{n}R2Eulers(\Delta R^*)\right) \tag{13}$$

$$\Delta t_i^* = \frac{i}{n}\Delta t^* \tag{14}$$

## 4. Experiments and Analysis

This method is designed for multi-beam LiDAR odometry, and the final result is evaluated in KITTI odometry benchmark. All the experiments, including the experiments of other methods, are based on the dataset from the benchmark.

KITTI odometry dataset is one of the most widely used datasets for evaluating computer vision algorithms in autonomous driving scenarios. Its LiDAR data is collected with Velodyne HDL-64E, which has 64 laser beams. The dataset has 22 sequences covering many different scenes, such as city streets, high way, etc.. The first 11 sequences contain ground-truth poses for all the data frames. They are used for the experiment of comparison with other methods and the odometry refinement experiment for our own method.

Two additional points should be noticed about the dataset. Firstly, the data is collected during the movement of the car, so the points with different yaw angles may be collected in different locations. However, the dataset provides the corrected data using the built-in high accuracy position and pose measurement sensors. Secondly, however, according to the work in [46], there is an error of 0.22° in the Velodyne sensor. We verified it by ourselves using the data in KITTI and applied it for our experiments.

### 4.1 Implementation Details

#### 4.1.1 Spherical Ring Projection

Velodyne HDL-64E is a 3D 64-beam laser scanner with a 10Hz rotation frequency. It has a 360° horizontal scanning range with 0.18° angular resolution and a $[-24.8°, 2°]$ vertical scanning range with a proximate 0.4254° angular step between laser beams. The distance measurement accuracy is 2cm, and it collects ~ 1.3 million points/second.

The projection parameters of spherical ring are set according to the scanning geometry. But the actual project spherical ring image has many hollow pixels if the scanning resolution is followed exactly. This is mainly due to the movement during the collection. To avoid this properly, we set the projection angle resolution smaller than the real projection angle resolution. But no more options such as noise filtering are used, trying to reduce the work on handcrafted algorithms and manually adjusted parameters. Even so, there are still some hollow pixels in spherical rings, but we leave this to be solved by CNN automatically. In the experiment, we set $\Delta\alpha = 0.2°$ and $\Delta\beta = 0.4254°$.

Still, because of the platform movement during the data collection, some points may exceed the vertical angle range of $[-24.8°, 2°]$. To save as much data as possible, we set the number of rows in spherical ring model as 69, not as the original number 64. Besides, there will be more than one point in a same spherical ring pixel during the projection, the last one dropped into the pixel will cover the previous ones.

#### 4.1.2 Voxel Model

There are several ways to extract multi-scale voxel patches around detected interest points. For example, the multi-scale voxel model can be projected and saved in local files at first, and then the voxel patches can be extracted according to the voxel indexes computed from the point coordinate values. An alternative way is to get the neighbor points by kNN algorithm [47] at first and then the voxel patches can be obtained by the voxelization of the neighbor points with different voxel resolutions. Here we take a compound method from these two methods. We save all the valid voxel indexes of the multi-scale voxel models during the pre-process, and the voxel patches in multiple scales are got by neighbor voxels searching using kNN algorithm firstly and by filtering out the voxels outside of the patches secondly. The parameters for the voxelization are mentioned in Section 3.3, as $S = \{0.02m, 0.16m, 0.64m\}$ and $S_P = 16$.

#### 4.1.3 Network Details

The details of the networks in Section 3.2 and in Section 3.3 are shown in Table 1 and Table 2 respectively (the number of filters and the number of channels are shown in bold fonts). Although both of the networks are convolutional auto-encoders, they are different in many aspects. Besides that, the first network is a 2D CAE and another one is a 3D CAE, the biggest difference is that the former one uses only convolutional layers as a local feature response while the latter one uses both convolutional layers and fully connected layers for extracting a 1D vector as the feature description of the input. The activation functions and loss functions used in the networks are different too. Firstly, we mostly use *relu* activation function for both of the networks except some special layers. In the 2D CAE, only the activation function of the output layer is set as *linear* because the output is forced to be the same as the input which contains negative values. As for the 3D CAE, we set *linear* activation function in the middle layer which is the feature layer and set *sigmoid* activation function in the last layer which outputs values of 0 and 1. Secondly, because the values in the input and the output are just coordinate values, the loss function in the 2D CAE is set to *mean squared error*, while the loss function in the 3D CAE is set to *binary cross-entropy*.

**Table 1.** Network details of the 2D CAE.

| Layer | Kernel/Pool size | Output size | Activation |
|---|---|---|---|
| Conv. | $3 \times 3 \times 3, \mathbf{32}$ | $64 \times 1792 \times \mathbf{32}$ | ReLU |
| Conv. | $1 \times 1 \times 32, \mathbf{8}$ | $64 \times 1792 \times \mathbf{8}$ | ReLU |
| Max. | $2 \times 2 \times 1$ | $32 \times 896 \times \mathbf{8}$ | - |
| Conv. | $3 \times 3 \times 8, \mathbf{16}$ | $32 \times 896 \times \mathbf{16}$ | ReLU |
| Max. | $2 \times 2 \times 1$ | $16 \times 448 \times \mathbf{16}$ | - |
| Conv. | $3 \times 3 \times 16, \mathbf{16}$ | $16 \times 448 \times \mathbf{16}$ | ReLU |
| Up. | $2 \times 2 \times 1$ | $32 \times 896 \times \mathbf{16}$ | - |
| Conv. | $3 \times 3 \times 16, \mathbf{8}$ | $32 \times 896 \times \mathbf{8}$ | ReLU |
| Up. | $2 \times 2 \times 1$ | $64 \times 1792 \times \mathbf{8}$ | - |
| Conv. | $1 \times 1 \times 8, \mathbf{3}$ | $64 \times 1792 \times \mathbf{3}$ | Linear |

**Table 2.** Network details of the 3D CAE.

| Layer | Kernel/Pool size | Output size | Activation |
|---|---|---|---|
| Conv. | $3 \times 3 \times 3 \times 1, \mathbf{8}$ | $16 \times 16 \times 16 \times \mathbf{8}$ | ReLU |
| Max. | $2 \times 2 \times 2 \times 1$ | $8 \times 8 \times 8 \times \mathbf{8}$ | - |
| Conv. | $3 \times 3 \times 3 \times 8, \mathbf{16}$ | $8 \times 8 \times 8 \times \mathbf{16}$ | ReLU |
| Max. | $2 \times 2 \times 2 \times 1$ | $4 \times 4 \times 4 \times \mathbf{16}$ | - |
| Conv. | $3 \times 3 \times 3 \times 16, \mathbf{32}$ | $4 \times 4 \times 4 \times \mathbf{32}$ | ReLU |
| Flatten | - | 2048 | - |
| Dense | - | 200 | ReLU |
| Dense | - | 20 | Linear |
| Dense | - | 200 | ReLU |
| Dense | - | 2048 | ReLU |
| Reshape | - | $4 \times 4 \times 4 \times \mathbf{32}$ | - |
| Conv. | $3 \times 3 \times 3 \times 32, \mathbf{16}$ | $4 \times 4 \times 4 \times \mathbf{16}$ | ReLU |
| Up. | $2 \times 2 \times 2 \times 1$ | $8 \times 8 \times 8 \times \mathbf{16}$ | - |
| Conv. | $3 \times 3 \times 3 \times 16, \mathbf{8}$ | $8 \times 8 \times 8 \times \mathbf{8}$ | ReLU |
| Up. | $2 \times 2 \times 2 \times 1$ | $16 \times 16 \times 16 \times \mathbf{8}$ | - |
| Conv. | $3 \times 3 \times 3 \times 8, \mathbf{1}$ | $16 \times 16 \times 16 \times \mathbf{1}$ | Sigmoid |

All the LiDAR data in KITTI benchmark are used for unsupervised training. The training of the networks is done on two NVIDIA 1080ti graphics cards, with 10 epochs for 2D CAE and 10 epochs for 3D CAE.

**4.2 Frame-to-Frame Matching**

We evaluate our method and compare our method with 3DFeatNet and USIP directly based on the frame to frame matching results on KITTI dataset with the same key parameters. The maximum number of interest points in these three methods is set to 1024. At the same time, we use RANSAC to do the frame to frame feature-based matching and set the inlier threshold to 1.0m, the minimum iterations to 100, and the maximum iterations to 10000. Additionally, for the parameters in our method, we set the nearest distance for interest points $\sigma$ to 10m and set the threshold $\delta$ of filtering out mediocre points to 0.2.

The experiments of frame to frame matching are based on the first 11 sequences in KITTI benchmark dataset. The performance is evaluated by Relative Translation Error (RTE), Relative Rotation Error (RRE), success rate, inlier ratio of the feature matching, and the average iterations during the RANSAC. Among them, the computation of RTE and RRE follows the work in [48]. A matching is regarded as success if $RTE < 0.5m$ and $RRE < 1°$.

Each method contains two parts: interest point detection and feature extraction. Each part makes efforts to the final feature-based matching result. To decouple the contributions of the two parts and to highlight the characteristics of the two parts, we make three tables, Table 3-1, Table 3-2, and Table 3-3, with different comparison principles. The data in the three tables are from the same source, but the first two tables sort the results in different ways to show different comparisons, and the third table only compares the results of the three whole methods.

In Table 3-1, the sorted three groups of results show the comparison of interest point detection. One conclusion can be observed: our interest points have the best performance on RTE and success rate, while USIP interest points have the best performance on RRE, inlier ratio, and average iterations. This difference can be explained in three aspects.

1) Multi-beam LiDAR collects points with fixed angular resolutions. However, USIP detects interest points by estimating their locations not by picking from point cloud. This gives USIP a smaller RRE. The higher inlier ratios and higher average iterations mean better repeatability of USIP interest points.

2) The interest points detected by USIP are always concentrated on structured locations with a relatively bigger scale, so as to USIP is more likely to ignore the interest points in details. Therefore, in some scenarios that are not that structured, USIP interest point tends to cause matching failure. On the contrary, our method can sensitively detect interest points in local details (see the comparison in Figure 7), so that in many scenarios where USIP interest point fails, it can successfully detect matching points, thereby having a higher success rate.

3) Because of the similar reason in 2), our method has a lower RTE. The experimented numbers of interest points in the work of USIP are 128, 256, and 512 [8]. For a well-performed comparison, we set the number of interest points in USIP to 1024, and the authors provided their trained model to us. According to our experiment, USIP also improved the accuracy of matching after increasing the number of interest points from 256 to 512 and 1024. However, USIP doesn't increase its attention on local details because of the increase in the number of interest points as its interest points usually appear repeatedly in the same places, while our method can be more sensitive to local details. Therefore, from this perspective, our method can detect more interest points with more details and thus has a lower RTE.

The matching cases in Figure 6 and Figure 7 can explain the reasons described in 2) and 3) well. Both methods can successfully match in the first matching case. But in the second matching case, which is a less structured scenario, the only way to match it successfully is to at least find out the interest points in small scales. Apparently, USIP cannot detect out enough suitable interest points, which causes the matching to fail. However, our method can detect the small corners on both sides of the highway, thereby significantly improving the matching success rate in these scenarios. From another perspective, although the numbers of interest points are the same in the two methods, it still looks like our method detects more interest points. That is because of that in the case of 1024 interest points, USIP interest points are concentrated in the same places repeatedly or concentrated on the areas of intersection of the ground and the walls. This phenomenon can also be found in the second matching case. Therefore, the matching performance cannot be effectively improved.

From Table 3-2, we can get a comparison conclusion: our descriptor can provide the best performance on most of the metrics: RTE, RRE, inlier ratios, and average iterations. This is mainly due to the fact that our feature descriptor can provide robust and reliable feature descriptions on three scales as small as 0.32m and as big as 10.24m. Small scale feature description can sharply capture local details while middle-scale feature description and big-scale feature description can make sure to get a high matching accuracy rate. Hence, our features can significantly improve success rate on finding matching pairs for RANSAC to have an over 50% higher inlier ratio ( $((((63.6 - 41.3))/41.3 + ((80.4 - 53.1))/53.1 + ((65.7 - 42.1))/42.1))/3 = 53.82\%$) compared to other methods. Thereby, the average iterations, RTE and RRE can be naturally reduced. As for the success rate, there is no conclusion that one of the descriptors can provide the best performance on this performance indicator. Because success rate is more sensitive to the quality of the interest point detection, at least based on these three descriptors.

Table 3-3 shows the comparison of the results from three full methods during the frame to frame matching. From the performance comparison, our method gets the smallest RTE and USIP gets the smallest RRE. Our method has significant advantages in success rate, inlier ratio, and average iterations. This also matches the conclusions and analysis from Table 3-1 and Table 3-2.

In summary, our interest points can capture more local details so that it can still find enough matching points in scenes where the structures are not obvious, and our features reach the state-of-the-art on the inlier ratio and the average iterations during feature-based matching.

**Table 3-1.** Frame-to-frame matching comparison for interest point detection on KITTI.

| **Interest point** + Desc. | RTE (m) | RRE (°) | Success Rate (%) | Inlier Ratio (%) | Avg # iter |
|---|---|---|---|---|---|
| 3DFeatNet + 3DFeatNet | 0.080±0.106 | 0.225±0.152 | 99.543 | 34.6 | 188.1 |
| USIP + 3DFeatNet | 0.065±0.137 | **0.153±0.118** | 99.047 | **49.3** | **125.6** |
| **Ours** + 3DFeatNet | **0.057±0.064** | 0.183±0.131 | **99.737** | 33.1 | 279.1 |
| 3DFeatNet + USIP | 0.096±0.136 | 0.239±0.164 | 99.306 | 41.3 | 141.3 |
| USIP + USIP | 0.065±0.116 | **0.167±0.116** | 99.526 | **53.1** | **119.2** |
| **Ours** + USIP | **0.065±0.072** | 0.199±0.129 | **99.746** | 42.1 | 211.6 |
| 3DFeatNet + Ours | 0.070±0.130 | 0.221±0.160 | 99.345 | 63.6 | 100.3 |
| USIP + Ours | 0.061±0.138 | **0.153±0.120** | 98.801 | **80.4** | **100.1** |
| **Ours** + Ours | **0.054±0.063** | 0.178±0.122 | **99.802** | 65.7 | 100.8 |

**Table 3-2.** Frame-to-frame matching comparison for feature description on KITTI.

| Interest point + **Desc.** | RTE (m) | RRE (°) | Success Rate (%) | Inlier Ratio (%) | Avg # iter |
|---|---|---|---|---|---|
| 3DFeatNet + 3DFeatNet | 0.080±0.106 | 0.225±0.152 | **99.543** | 34.6 | 188.1 |
| 3DFeatNet + USIP | 0.096±0.136 | 0.239±0.164 | 99.306 | 41.3 | 141.3 |
| 3DFeatNet + **Ours** | **0.070±0.130** | **0.221±0.160** | 99.345 | **63.6** | **100.3** |
| USIP + 3DFeatNet | 0.065±0.137 | 0.1532±0.118 | 99.047 | 49.3 | 125.6 |
| USIP + USIP | 0.065±0.116 | 0.167±0.116 | **99.526** | 53.1 | 119.1 |
| USIP + **Ours** | **0.061±0.138** | **0.1526±0.120** | 98.801 | **80.4** | **100.1** |
| Ours + 3DFeatNet | 0.057±0.064 | 0.183±0.131 | 99.737 | 33.1 | 279.1 |
| Ours + USIP | 0.065±0.072 | 0.199±0.129 | 99.746 | 42.1 | 211.6 |
| Ours + **Ours** | **0.054±0.063** | **0.178±0.122** | **99.802** | **65.7** | **100.8** |

**Table 3-3.** Frame-to-frame matching comparison for full methods on KITTI.

| Interest point + Desc. | RTE (m) | RRE (°) | Success Rate (%) | Inlier Ratio (%) | Avg # iter |
|---|---|---|---|---|---|
| 3DFeatNet + 3DFeatNet | 0.080±0.106 | 0.225±0.153 | 99.543 | 34.6 | 188.1 |
| USIP + USIP | 0.065±0.116 | **0.167±0.116** | 99.526 | 53.1 | 119.2 |
| **Ours + Ours** | **0.054±0.063** | 0.178±0.122 | **99.802** | **65.7** | **100.8** |

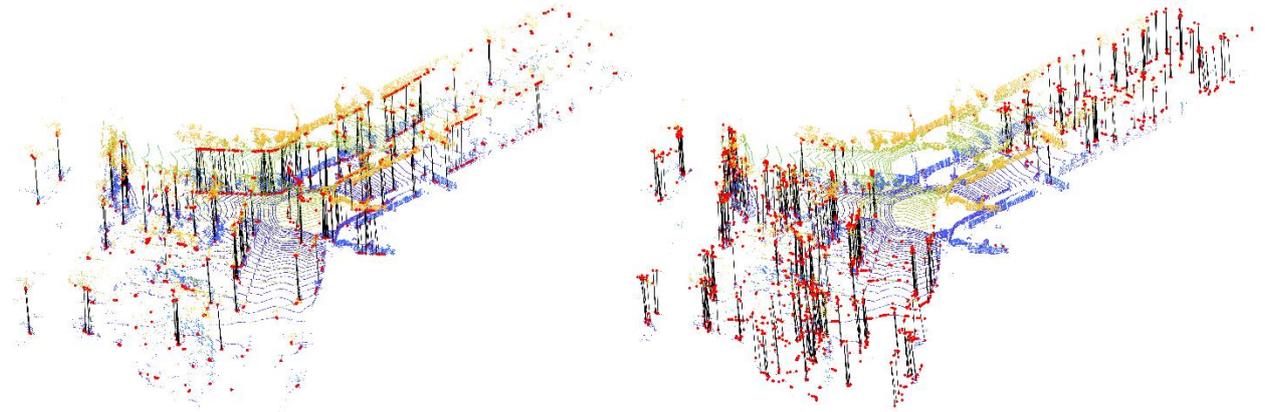

**Figure 6.** The comparison on interest detection and feature matching of the method USIP (left) and our method (right) in an ordinary matching scenario. The red points are the detected interest points and the black lines are the matches after RANSAC. Both methods detect 1024 interest points and take 100 iterations. USIP gets a 59.0% inlier ratio and our method gets a 62.1% inlier ratio.

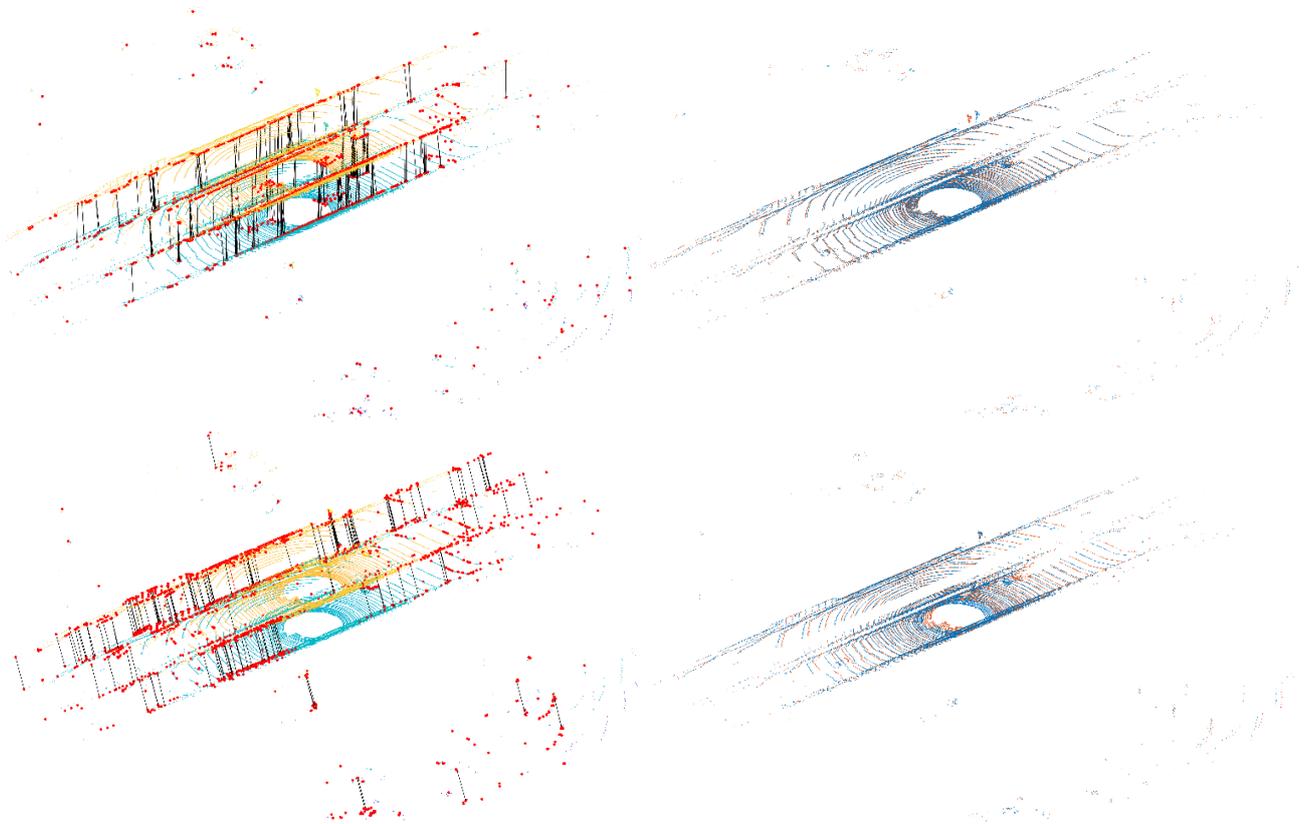

**Figure 7.** The matching comparison of the method USIP (first row) and our method (second row) in a difficult scenario. In the first column, the red points are the detected interest points and the linking lines are the matches after RANSAC. The second column is the fused point cloud after the matching, with different point clouds are shown in different colors. Both methods detect 1024 interest points. USIP takes 197 iterations with a 28.5% inlier ratio and our method takes 137 iterations with a 32.1% inlier ratio.

### 4.3 Odometry Refinement and Result

During the frame to frame matching, the indexes of the matching pairs are saved to local files. After that, the keyframes can be obtained according to the method in Section 3.4 by offline processing. Finally, the odometry refinement is done with the EIPs using ICP-based registration. In this experiment, for the extraction of EIPs, we set the parameter of $h_E$ to 7, and set the half neighbor size $h$ for EIPs to 2. To achieve higher accuracy, we also use the ground points with their normal vectors to join in the ICP-based registration. Since the obtaining of ground points and the normal vectors, not like the obtaining of EIPs, have little to do with the two CAEs, they are not further presented in this paper.

A statistic for the frame length between keyframes is made on the 11 sequences in KITTI, and the data distribution is shown in Figure 8. There are 2668 times of ICP-based refinements, most of the frame lengths are between 3 and 14. In other words, according to the data gathering frequency, the time difference between two keyframes is usually between 0.3 and 1.4 seconds. Therefore, it won't cause too much computation because of the too-short distance between keyframes and the registration failure due to the too long distance between keyframes can also be limited.

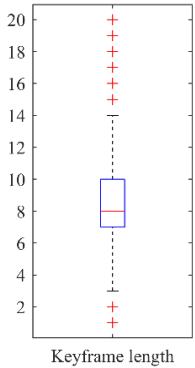

**Figure 8.** The distribution of the number of frames between two keyframes.

At the same time, we also do the statistic of the error decrease after the refinement. As shown in Figure 9, the RTE of the relative poses between keyframes decreases from $0.354m$ to $0.170m$. The RRE of the relative poses between keyframes decreases from $0.696°$ to $0.148°$. Also, the corresponding standard deviations decrease.

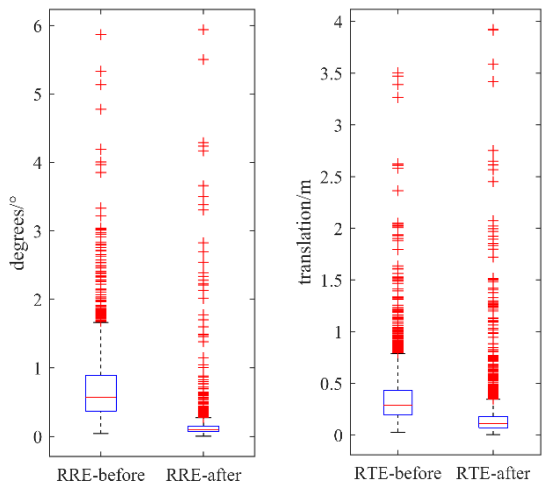

**Figure 9.** The distribution comparison of the relative pose error between two keyframes before (left) and after (after) the odometry refinement.

The backward pose update after the refinement for keyframes can eliminate the pose jump to make the trajectories be more real. The visualization of one case of backward pose update is shown in Figure 10. There will always be a pose jump before one keyframe, but after the backward pose update, the relative pose is smooth everywhere, including the directions of the frames.

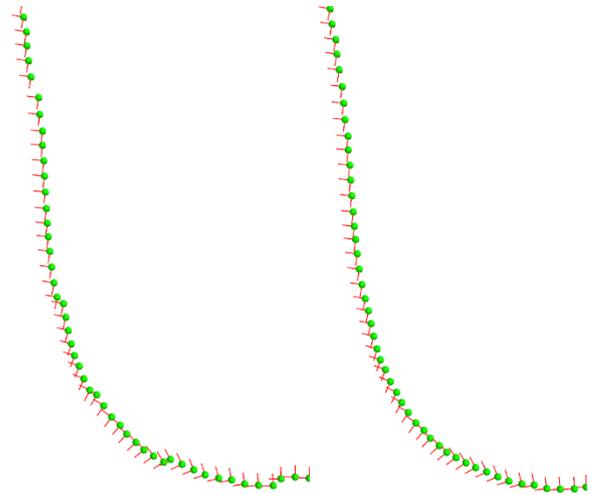

**Figure 10.** The comparison of odometry without and with backward pose update after refinement. Left, without backward update; right, with a backward update. Each green dot represents a position of a frame and the red arrows represent the axes of its coordinate system.

### 5. Discussion

The 2D CAE part utilizes the projected spherical ring data structure, which is a very compact data representation method for LiDAR point cloud and which also makes it very suitable for CNNs with mature technology. Firstly, EIPs, which are used for ICP-based registration refinement, can be easily extracted from spherical rings according to the detected interest points. Secondly, based on the projected spherical ring, further research aiming at other tasks such as segmentation can be conducted. Thirdly, the idea of interest point detection can also be used in any other 2D structured data like 2D ordinary images for image registration.

The 3D CAE part shows a solution for multi-scale feature extraction based on multi-resolution voxel model. The voxel model has the advantages of preserving the original data shape and the constant real-size scale in each location, which is very suitable for feature extraction. Otherwise, a bigger voxel patch with a small voxel size aiming at extracting higher location accuracy and a bigger receptive area will lead to a very high hardware consumption and a slow operation. The multi-resolution voxel model is similar to the idea of image pyramid, which can provide multi-scale features with the same neighbor data size. This point, combined with the idea of 3D CAE with unsupervised training, makes our descriptor reach over 50% improvement of the performance compared to the state-of-the-art.

There are some additional notes on the two parts. Firstly, our interest point is more likely to detect the sharp details, not like the detecting strategy in UISP. This difference can be found in Figure 6 and Figure 7. Secondly, our interest points are also scattered in distance areas more. According to the authors in [46], points in close distance are not very effective at constraining the rotations of a scan registration problem. So, from this perspective, our more scattered distance interest points let our method have advantages in feature-based registration. Thirdly, many methods based on PointNet, such as 3DFeatNet and UISP, use downsampling to decrease the number of points, but our spherical ring model and the multi-scale voxel model can largely save the original information to maximize the matching accuracy. Fourthly, there is no particular design for rotation invariant of the descriptor, which means our descriptors may fail to match if there is a large angle difference between two matching frames. This is because that CNN itself has some rotation invariance [49] and also because that the frame to frame matching doesn't need significant rotation invariance. If needed, there are several ways to do this, such as utilizing direction estimation and alignment.

Besides the two CAEs for interest point detection and feature extraction, we also propose a keyframe selection method and a backward pose update method. As far as we know, this is the first time of using the matching pairs transferring to do the keyframe selection and there is also no literature that shows the same backward pose update method based on our idea. More applications based on these two parts should be carried out.

We name our method as CAE-LO, and after adding the ground constraint based on normal estimation, which is not shown in this paper, our method achieved 0.86% accuracy in KITTI benchmark [1]. We also release our code of the initial odometry part and the data used in the comparisons.

## 6. Conclusion

CAE-LO is proposed, leveraging fully unsupervised convolutional auto-encoder for interest point detection and feature extraction from multi-beam LiDAR point cloud. From the designed comparison experiments with the methods of state-of-the-art, our interest point detection is more capable to detect local details, thereby improving the matching success rate in scenarios where the structure is not obvious, and our features show over 50% improvement in matching inlier ratio.

Besides, as the important parts in many solutions like odometry and SLAM, the proposed matching pair transferring based keyframe selection method, the ICP-based registration using EIPs, and the backward pose update method show their feasibility and accuracy improvement on refined odometry.

## Acknowledgments

This study was supported in part by the Natural Science Fund of China with Project No. 41874031, the Technology Innovation Program of Hubei Province with Project No. 2018AAA070, the Natural Science Fund of Hubei Province with Project No. 2018CFA007, China Scholarship Council (CSC) (No. 201806270196), MOE (Ministry of Education in China) Project of Humanities and Social Sciences with Project No. 18YJCZH242, and Humanities and Social Sciences project of Hubei Provincial Department of Education with Project No. 18Q059.